\documentclass[runningheads, envcountsame, a4paper]{llncs}
\usepackage[pdftex]{graphicx}
\usepackage{epstopdf}
\usepackage[hyphens]{url}
\usepackage{microtype}
\usepackage[ruled,vlined]{algorithm2e}
\usepackage{booktabs}
\usepackage{multicol}
\usepackage{multirow}
\usepackage{array}
\usepackage{floatrow}
\usepackage[misc]{ifsym}
\usepackage{titletoc}
\usepackage{amsfonts}
\newfloatcommand{capbtabbox}{table}[][\FBwidth]

\usepackage[table]{xcolor}% http://ctan.org/pkg/xcolor
\newcommand\MyBox[2]{
  \fbox{\lower0.75cm
    \vbox to 1.7cm{\vfil
      \hbox to 1.7cm{\hfil\parbox{1.4cm}{#1\\#2}\hfil}
      \vfil}%
  }%
}

\begin{document}

\title{Data-driven Policy on Feasibility Determination for the Train Shunting Problem}
\toctitle{Data-driven Policy on Feasibility Determination for the Train Shunting Problem}
\titlerunning{Data-driven Policy on Feasibility Determination for the TUSP}

\author{Paulo Roberto de Oliveira da Costa\inst{1} (\Letter) \and
Jason Rhuggenaath\inst{1} \and Yingqian Zhang\inst{1} \and Alp Akcay\inst{1} \and Wan-Jui Lee\inst{2} \and Uzay Kaymak\inst{1} }

\tocauthor{Paulo Roberto~de Oliveira da Costa, Jason~Rhuggenaath, Yingqian~Zhang, Alp~Akcay,   Wan-Jui~Lee, Uzay~Kaymak}

\authorrunning{P.R. de O. da Costa et al.}
\institute{Eindhoven University of Technology, 5612 AZ Eindhoven, Netherlands \and
Dutch Railways, 3511 CA Utrecht, Netherlands\\
\email{$\{$p.r.d.oliveira.da.costa, j.s.rhuggenaath, yqzhang, a.e.akcay, u.kaymak$\}$@tue.nl} \email{ wan-jui.lee@ns.nl}}
\maketitle              % typeset the header of the contribution
\begin{abstract}
Parking, matching, scheduling, and routing are common problems in train maintenance. In particular, train units are commonly maintained and cleaned at dedicated shunting yards. The planning problem that results from such situations is referred to as the Train Unit Shunting Problem (TUSP). This problem involves matching arriving train units to service tasks and determining the schedule for departing trains. The TUSP is an important problem as it is used to determine the capacity of shunting yards and arises as a sub-problem of more general scheduling and planning problems. In this paper, we consider the case of the Dutch Railways (NS) TUSP. As the TUSP is complex, NS currently uses a local search (LS) heuristic to determine if an instance of the TUSP has a feasible solution. Given the number of shunting yards and the size of the planning problems, improving the evaluation speed of the LS brings significant computational gain. In this work, we use a machine learning approach that complements the LS and accelerates the search process. We use a Deep Graph Convolutional Neural Network (DGCNN) model to predict the feasibility of solutions obtained during the run of the LS heuristic. We use this model to decide whether to continue or abort the search process. In this way, the computation time is used more efficiently as it is spent on instances that are more likely to be feasible. Using simulations based on real-life instances of the TUSP, we show how our approach improves upon the previous method on prediction accuracy and leads to computational gains for the decision-making process.
\keywords{Planning and scheduling \and Graph classification \and Local search \and Train shunting.}
\end{abstract}
\section{Introduction}

\begin{figure}[ht]
    \centering
    \resizebox{0.6\columnwidth}{!}{%
    \includegraphics{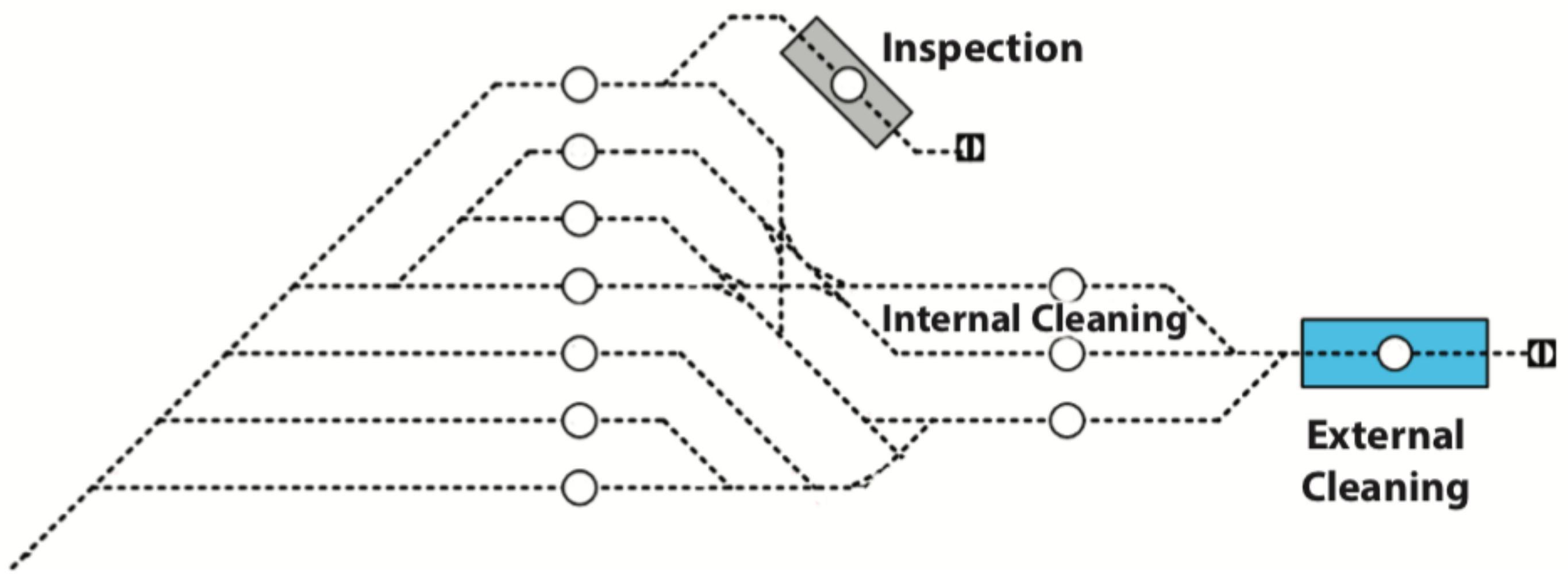}
    }
    \caption{ ``Kleine Binckhorst'' in The Hague. Shunting yard with specific tracks for inspection and cleaning tasks \cite{Spoorenplan}}
    \label{fig:service_site}
\end{figure}

\textit{Parking}, \textit{matching}, \textit{scheduling} of service tasks and \textit{routing} problems are common in railway networks and arise when trains are not in operation. In such cases, train units are maintained and cleaned at dedicated shunting yards (Figure \ref{fig:service_site}). The planning problem that arises from such situations is referred to as the Train Unit Shunting Problem (TUSP). The problem involves matching train units to arriving and departing train services as well as assigning the selected compositions to appropriate shunting yard tracks for maintenance.
To assess the capacity of each of its shunting yards, the Dutch Railways (NS) has developed an internal simulator. This simulator is used to both determine the capacity of shunting yards as well as analyse different planning scenarios. Currently, a Local Search heuristic (LS) applying a simulated annealing algorithm \cite{Vandenbroek2016} 
is used to find feasible solutions. The LS requires an initial solution as a starting point, and at the end of a run, the LS either returns a feasible or infeasible plan. The LS is more computationally efficient than the previously formulated mathematical optimisation model \cite{kroon2008shunting}. However, given the number of shunting yards and scenarios, the capability of improving the evaluation speed of the LS can bring significant computational gain to NS. 

In recent years, many studies have investigated using machine learning models to accelerate the search for optimal solutions when solving combinatorial optimisation problems  \cite{Khalil_NIPS2017_7214,Lombardi2018,verwer2017auction}. In the context of train shunting, the authors of \cite{peer2018shunting} consider the parking of trains as a complete (reinforcement) learning problem. In \cite{Dai2018,arno}, machine learning methods are used to learn the relationship between initial solutions an a feasibility output from the LS heuristic.
\begin{figure}[ht]
    \centering
    \resizebox{0.9\columnwidth}{!}{%
    \includegraphics{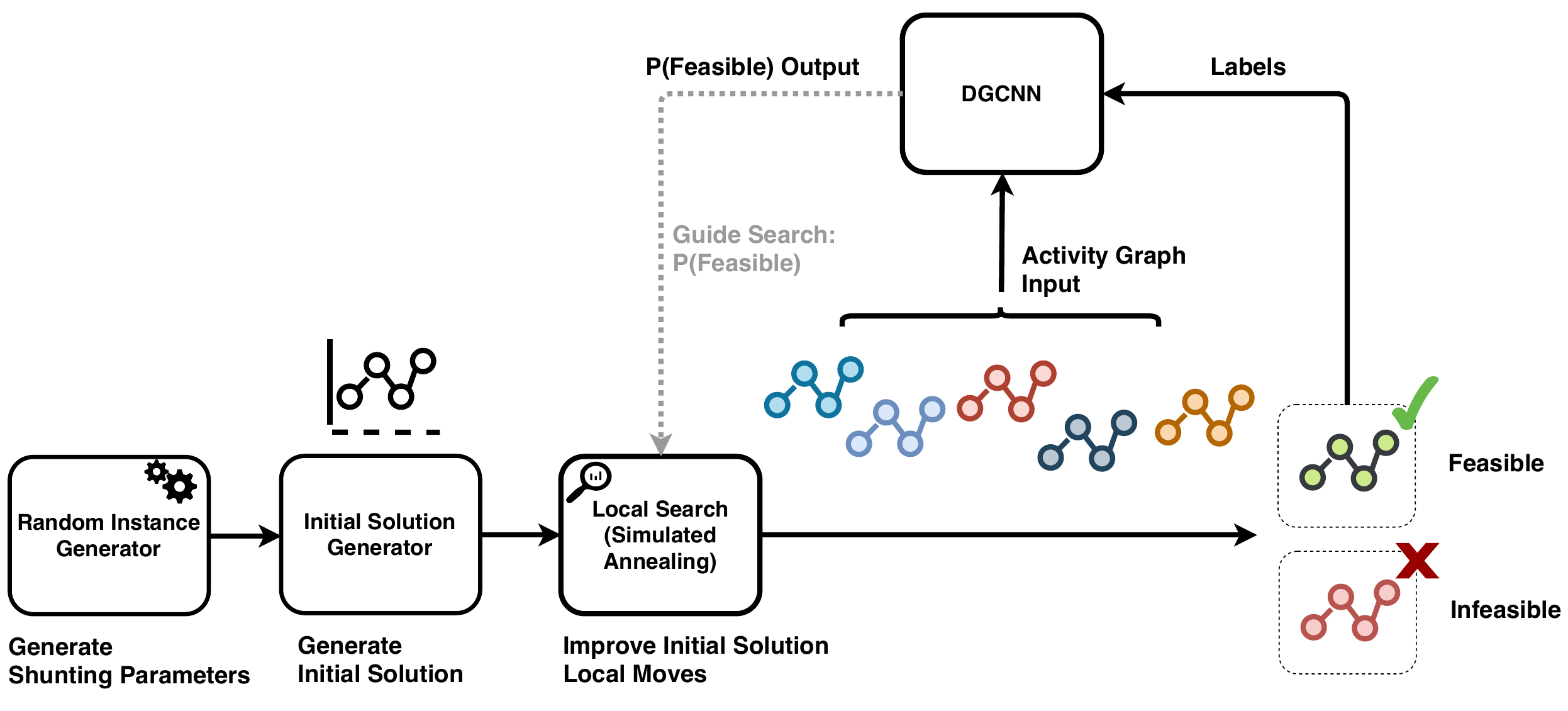}
    }
    \caption{Diagram depicting the proposed methodology}
    \label{fig:diagram}
\end{figure}
%
%
% In the context of train shunting, some works have focused on parts of the optimisation problem, for example, considering the parking of trains as a complete (reinforcement) learning problem \cite{peer}. However, no end-to-end learning approach exists for the complete TUSP formulation. More recently, \cite{arno} and \cite{Dai2018} have shown that machine learning methods can be used to learn the relationship between initial solutions an a feasibility output from the LS heuristic. 
%
% Given an initial LS solution, a classification model predicts whether the LS can find a feasible solution before applying the LS operators. 

In this paper, we use graph encoding \cite{Vandenbroek2016} to represent each intermediate solution as an activity graph of maintenance activities. Activity graphs allow us to use a graph representation of the solution space and search for local discriminative features of shunting plans. We then use a Deep Graph Convolutional Neural Network (DGCNN) \cite{Zhang2018} as a feature extractor to train a model that predicts the future feasibility of each precedence graph. We formalise the approach of \cite{arno} by including a sequence of seen shunting plans during an LS run as training examples. We show how to combine the predictions with the LS to derive a policy that decides to terminate the LS run based on the sequence of intermediate solutions seen so far. This way, the train shunting simulator can decide on whether or not it should invest time in a set of solutions alongside the LS, and can determine the feasibility of given instances with higher confidence. We present a schematic view of our approach in Figure \ref{fig:diagram}, and summarise or main contributions as follows:
\begin{itemize}
\item We demonstrate that encoding both activity graphs of (intermediate) solutions and important domain knowledge such as time-related information (see section \ref{subsec: graphconvshunt}) leads to better predictions on feasibility determination of planning instances. 
\item  We develop a learning policy that can be used along with local search on determining feasibility of a given planning instance. We show that taking into account the sequence of intermediate solutions in the search increases the prediction accuracy. 
\end{itemize}
The rest of our paper is organised as follows. In Section \ref{sec:back}, we describe the background information and related work. In Section \ref{sec:methodology}, we present the proposed algorithm and show how we learn the LS outputs using the proposed DGCNN. In Section \ref{sec:results}, we describe the experimental setup and the main results in comparison to the previously proposed method.

\section{Background and Related Work}
\label{sec:back}
\subsection{Train Unit Shunting Problem}

Shunting yards have some specific characteristics that make the shunting problem complex and interconnected. For example, routing movements can only happen over tracks, which imposes restrictions of possible movements and turns. Furthermore, some tracks can be approached from both ends while others can only be approached from one side of a track. Also, multiple trains can be parked on the same one-end track; therefore, trains have to be parked in the order they must leave as overtaking is not possible, while specific tracks exist for cleaning and inspecting. Coupling and decoupling of train units are also important. Train units can be combined to form longer compositions, units can be of different types, but only train units with
the same type can be combined. Lastly, trains have to leave at specific times to meet the transportation demands and ideally should not be delayed as this impacts the train network. 

The TUSP has been shown to be an NP-complete problem \cite{freling2005shunting,haahr2017integrating}. Several works have attempted to solve variations of the TUSP, including mixed integer programming (MIP), dynamic programming and column generation methods \cite{haahr2017optimization}. We focus our attention on the LS proposed by \cite{Vandenbroek2016} where a simulated annealing heuristic is proposed for finding shunting plans with service activities. The shunting plans include the \textit{matching} of trains, \textit{scheduling} of service tasks, assignment to \textit{parking} tracks and \textit{routing}. Different from other exact approaches, the algorithm integrates all the components of the TUSP simultaneously rather than sequentially. The LS heuristic has shown better performance on real-world instances when compared to a MIP formulation \cite{Vandenbroek2016} as well as being able to handle larger instances. More important, NS currently uses this heuristic in its Shunt Plan Simulator to define the capacity of its service sites.

The Shunt Plan Simulator at NS consists of three sequential stages: (1) instance generation, (2) initial solution generation, and (3) finding feasible solutions using the LS. The maximum capacity of a given shunting yard is then determined by repeatedly running the local search heuristic with different instances and scenarios. After a number of runs, the simulation converges towards a number of train units for which the heuristic can solve at least 95\% of the instances. The capacity of a shunting yard is defined as the number of train units it can serve during a 24-hour period. 

The instance generator derives instances for the TUSP automatically. Instances can be generated for each shunting yard individually based on a day schedule. Examples of parameters are number of train units, arrival/departure distributions and service tasks. The output of the instance generator is a set of arriving trains (AT), a set of departing trains (DT) and a set of service tasks (ST) for each train unit. For both AT and DT, train compositions, train units and arrival/departure time are specified. For each train unit, a list of ST is specified for the time they are present on the service site. Trains can be composed of one or more train units of the same type, which are a set of carriages that form a self-propelling vehicle. The same unit type can have multiple sub-types, where the sub-type indicates the number of carriages.
%
% Fig. \ref{fig:virms} shows a train unit type and its corresponding sub-types. 

% \begin{figure}[ht]
%     \centering
%     \resizebox{1\columnwidth}{!}{%
%     \includegraphics{figures/virms.pdf}
%     }
%     \caption{A train unit type with two sub-types: 6 and 4 carriages.}
%     \label{fig:virms}
% \end{figure}
%
The output of the instance generator serves as input for the initial solution generator. The algorithm of Hopcroft-Karp \cite{hopcroft1973n} is used to produce a matching between arriving and departing train units. Next, a service task schedule is greedily constructed to form an initial solution. Note that, in general, initial solutions are not feasible, as they may violate temporal or routing constraints.
% An initial solution contain all important features to act as a starting point for the LS heuristic. 
After the initial solution is found, the LS applies 11 local search operators to move through the search space. Intermediate random restarts are used if no improvement can be found for a specified running time. The LS ends when a feasible solution has been found or when a maximum running time is reached. A detailed explanation of the heuristic can be found in \cite{Vandenbroek2016}.
% However, it could be expanded to search for feasible solutions with lower objective costs. 
% Experiments showed that the LS is capable to find feasible shunt plans in both artificial and real-world scenarios \cite{Vandenbroek2018}. The performance of the LS has been compared to a mathematical optimisation model. The results showed that the LS heuristic was capable of finding the TUSP planning for more train units in most experiments. A detailed explanation of the heuristic can be found in \cite{Vandenbroek2018,Vandenbroek2016}. 
\subsection{Graph Classification}

\begin{figure}[ht]
    \centering
    \resizebox{0.9\columnwidth}{!}{%
    \includegraphics{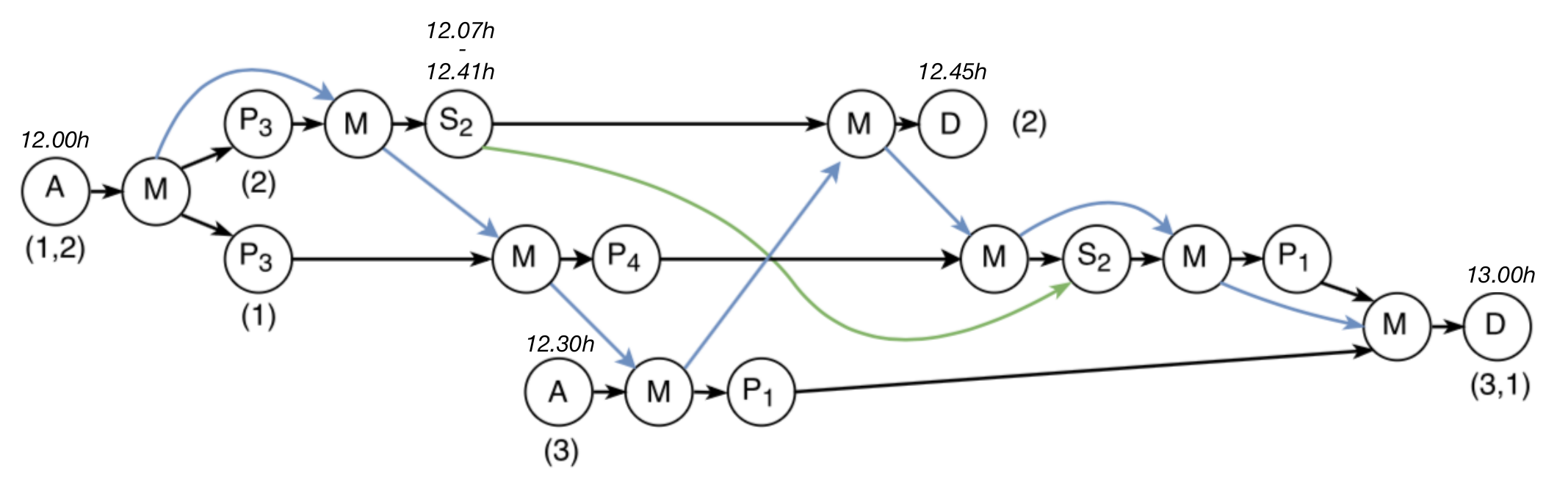}
    }
    \caption{The activity graph of a shunting plan. A train unit number is encoded in brackets. Activity nodes are encoded with starting and ending times. Black edges represent the order of operations of train units. Blue edges represent the order of the movements and green edges indicate which service task is completed first. Activity nodes encode specific service tasks and parking tracks as a subscript \cite{arno}}
    \label{fig:activity_graph}
\end{figure}

A shunting plan can be modelled as an \textit{activity graph}. An activity graph contains all the activities that have to be completed during a plan. Moreover, it represents the dependencies and activity order via a directed graph. Figure \ref{fig:activity_graph} depicts an activity graph of a shunting plan. The activities nodes, including Arrival (A), Service (S), Parking (P), Movement (M) and Departure (D), are connected by edges indicating the precedence relationships. Corresponding train units are assigned to activities nodes representing the complete shunting plan. Given the graph structure that arises from shunting plans, we relate our problem setup to that of graph classification. 
% The goal of a graph classification task is to learn a function that can map arbitrary graph structures to designated labels (classes). 

Previously proposed methods in graph classification can be subdivided into graph kernel methods and topological methods. Graph kernels measure the similarity between graphs by directly comparing substructures within graphs. For example,  \cite{shervashidze2011weisfeiler} presented a family of graphs kernels using the  Weisfeiler-Lehman (WL) test of isomophormism to extract graph features. Results show that the features can capture node and topological information and are competitive in comparison to other graph classification methods. On the other hand, topological methods extract features directly from graphs. Such features can represent either local (e.g. node degrees) or global (e.g. number of nodes) information about an input graph. The extracted features can be combined to create a multidimensional input vector to a machine learning algorithm \cite{bonner2016deep,akoglu2015graph,aggarwal2014evolutionary}.  
%
% Even though it can require substantial computational time to create useful features coming from large graphs, 
% Topological methods have been successfully applied to graph mining tasks, e.g. graph similarity measurement \cite{bonner2016deep}, time series anomaly detection \cite{akoglu2015graph} and link prediction \cite{aggarwal2014evolutionary}. 
%
More recently, methods that can extract useful features directly from graph representations without computing graph kernels or topological features have been proposed. Current state-of-the-art results have been achieved using Convolutional Neural Networks (CNN) tailored to extract graphs features automatically. Such methods showed competitive performance against other graph kernel algorithms \cite{Zhang2018,Niepert2016,Kipf2016}.

For the TUSP, a topological method \cite{Dai2018} has been proposed to extract local and global features from initial solutions. The features are then used to predict the feasibility of an initial solution using several machine learning algorithms. Results show that time-related features are the most import for prediction accuracy in the tested data. Later, \cite{arno} uses a DGCNN \cite{Zhang2018} to extract graph features (node classes) directly from the activity graphs of initial solutions without incurring in manual feature engineering. The results show that the DGCNN can achieve similar performance when compared to \cite{Dai2018}, suggesting that the DGCNN model is able to extract useful features from shunting plans.

Our work builds upon the work of \cite{arno} as to generalise the feasibility prediction to an arbitrary graph during the LS run. We improve on the original model and use the DGCNN to extract node and time features (see section \ref{subsec: graphconvshunt}) from the shunting plans. 
% Moreover, we show that using the graphs generated by LS procedure and the initial solutions, yield better results for the initial solution prediction problem. 
Lastly, we provide a generalisation of the algorithm and show that we can use it to predict the feasibility of a given plan in a local search run at each iteration. This modification allows for more saved time as a prediction can be made at each iteration and computation can be halted on unpromising search space.

\section{Methodology}
\label{sec:methodology}
\subsection{Deep Graph Convolutional Neural Network}

A Deep Graph Convolutional Neural Network (DGCNN) \cite{Zhang2018} accepts graphs of arbitrary structure as inputs. It aims at extracting useful deep features characterising the information encoded in a graph and determining how to sequentially read a graph in a meaningful and consistent order. Graph convolution layers are used to extract local substructure features and define a consistent node ordering. The convolution layers mimic the behaviour of the  Weisfeiler-Lehman Subtree Kernel \cite{shervashidze2011weisfeiler} and the Propagation Kernel \cite{neumann2016propagation} which are commonly used to extract graph features in classification tasks. To sequentially read graphs in a consistent order, a \textit{SortPooling} layer sorts the nodes under a predefined order and unifies input sizes. The layer achieves a fixed length representation of a graph that is later combined with traditional convolutional and dense layers to map the inputs to an output class. Empirical results have shown that the DGCNN achieved state-of-the-art performance  on  several graphs classification tasks. In our work, we use a slightly modified version of the DGCNN to extract features from shunting plans.
\begin{figure}[!ht]
%  \vspace{-0.2cm}
  \centering
      \resizebox{1.0\columnwidth}{!}{%
    \includegraphics{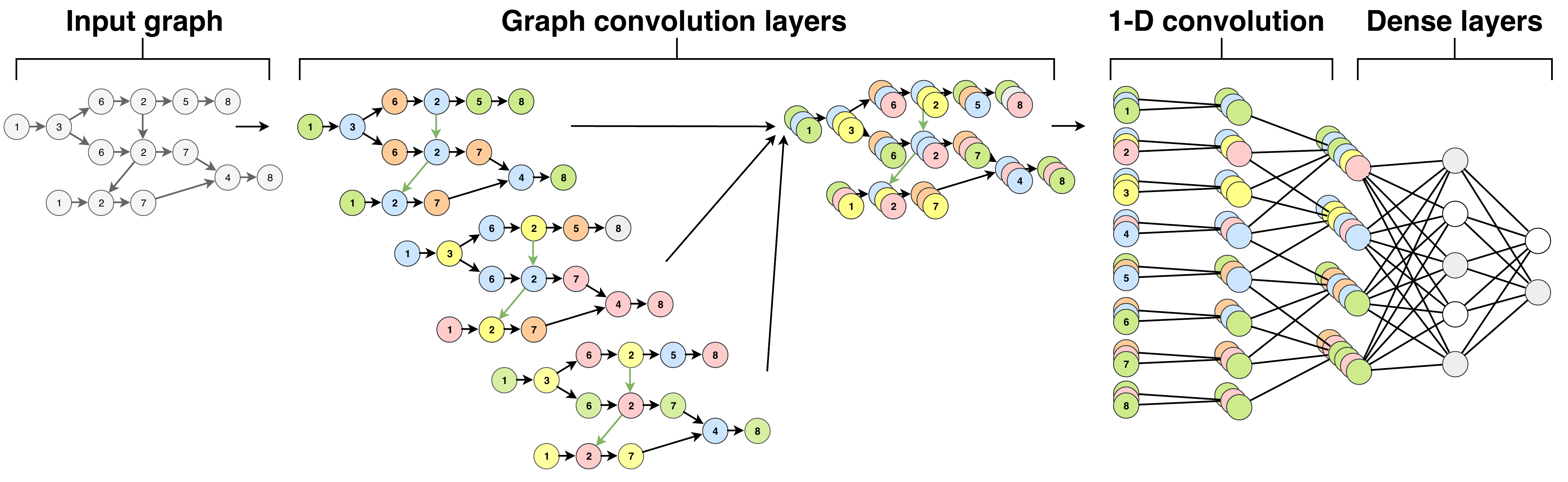}
    }
  \caption{The DGCNN architecture for shunting plans adapted from \cite{Zhang2018}. An input graph of arbitrary structure is first passed through multiple graph convolution layers where node labels are propagated between neighbours, visualised as different colours. Node features are then passed to traditional CNN structures \cite{arno}}
  \label{fig:DGCNN}
\end{figure}
\subsection{Graph Convolution of Shunting Plans}
\label{subsec: graphconvshunt}

We denote a graph representation of a shunting plan as a graph $G$ represented by $G = (V,E)$ where $V$ represents represents the set of nodes and $E$ represents the set of edges on the graph. 
%Edges are represented by a tuple connecting two nodes $\{u, v\}$ where $u, v \in V$. %
We use $n$ to determine the total number of nodes in a graph. Moreover, our nodes encode eight different activities in a shunting plan, these activities are represented as node labels. The original representation is modified to include more information about the graphs. We include specific types of activities to effectively exploit the second localized graph convolution of the DGCNN which involves appending node labels of neighbouring nodes.

Our shunting plans contain, among others, Parking ($P$) and Service ($S$) activity nodes. Similar to \cite{arno}, we encode information on $P$ and $S$ activities by appending extra information to the nodes. The specific parking and services tracks are appended to the respective nodes to encode more information about the moves.  
%
% The specific parking and services track is appended to parking nodes $P_i$, where $i = 1,...,P_T$ and $P_T$ is the number of parking tracks on a shunting yard. The specific service task is appended to get $S_i$, where $i = 1,..., S_T $ and $S_T$ is the number of service tasks that can be performed on a shunting yard. 
% Previous results \cite{arno} showed that including both more node labels improves classification performance. 
%
In the topological model of \cite{Dai2018}, temporal features are deemed important to extract useful information from scheduling plans. In particular, the time between train arrivals and service tasks are the most important features to predict feasibility, as shown in \cite{Dai2018}. Therefore, to aggregate more temporal information from time features for each activity node in a plan, we also encode the following time-related features:  

\begin{itemize}
    \item $x_{s}  \in \mathbb{R}^{n}$: \textit{time between the current activity and the start of the plan}.
    \item $x_e \in \mathbb{R}^{n}$: \textit{time between the current activity and the end of the plan}.
    \item $x_d  \in \mathbb{R}^{n}$: \textit{activity duration}. 
    \item $x_a \in \mathbb{R}^{n}$: \textit{average time between an activity and all its adjacent activity nodes}.
\end{itemize}
% \textit{time between the current activity and the start of the plan}, the \textit{time between the current activity and the end of the plan}, \textit{the activity duration} and \textit{the average time between an activity and all its adjacent activity nodes}.

We denote a 0/1 symmetric adjacency matrix of a graph as $A$. We use $X \in \mathbb{R}^{n \times c}$ to denote a graph's node information matrix with each row representing a node and a $c$-dimensional feature vector. Our features correspond to both node labels and to time features. As proposed by \cite{Zhang2018} we stack multiple graph convolutions of the form:
\begin{equation}\label{eq:graph_conv}
    Z^{t+1} =   f(\tilde{D}^{-1}\tilde{A}Z^{t}W^{t})  
\end{equation}
where $Z^{0} = X$,  $\tilde{A} = A + I$ is the adjacency matrix of the graph with self-loops, $\tilde{D}$ is its diagonal degree matrix with $\tilde{D}_{ii} = \sum_j \tilde{A}_{ij}$, $W^t \in \mathbb{R}^{c_t \times c_{t+1}}$ is a matrix of trainable convolution parameters and $c_t$ are the convolution channels.  $f$ is a nonlinear activation function, $Z^t \in \mathbb{R}^{n \times c_t} $  is the output activation matrix and $t$ represents the $t$-th convolution layer. After multiple graph convolution layers (Eqn. \ref{eq:graph_conv}), the outputs are combined to generate multiple feature channels for a given shunting plan graph. We concatenate the output of $h$ graph convolution layers horizontally to form a concatenated output, written as $Z^{1:h} \in \mathbb{R}^{n \times \sum_{1}^h c_t}$. Unlike \cite{Zhang2018} we do not implement the \textit{SortPooling} layer as our graphs already represent an ordered shunting plan. However, we unify the output tensor in the first dimension from $n$ to $k$. After, several 1-D convolution and MaxPooling layers are concatenated to a dense layer and an output layer to form the output. A chart view of the graph convolutions is shown in Figure \ref{fig:DGCNN}.

\subsection{Learning Feasibility using Local Search}

To learn the future feasibility of a given shunting plan, we generate instances with a number of trains $\tau$, and run the LS procedure for each instance for a predefined maximum running time $r$. We denote as $L$ the total number of runs of the LS procedure where $l$ represents the $l$-th LS run for $l = 1,..., L$. Thus, $L$ is a parameter of the LS and has to be selected depending on the problem instance. For each LS run, we generate graphs $G_i$ where $i = 0,..,N_l$, where $N_l$ is the total number of graphs for the $l$-th LS run and $G_0$ is the initial solution. 

For each run of the LS an associated class can be retrieved at the end of the procedure representing either a feasible ($1$) or infeasible solution ($0$). Moreover, each graph $G_i$ can have its corresponding class $y_i \in \{0,1\} $ according to the number of look-ahead iterations $W$ considered. That is, after each LS run, a graph $G_i$ in the sequence of solutions of the LS will have $y_i$ defined by the feasibility of the solution $G_{min\{N_l, W+i\}}$. The parameter $W$ is a hyper-parameter of the proposed method, and when $W \geq N_l$ all the graphs in a run will have as output class the feasibility of the final solution of the LS.
\begin{algorithm}[ht]
\SetKwFunction{LS}{LS}
\SetKwFunction{DGCNN}{DGCNN}
\SetKwFunction{TrainDGCNN}{TrainDGCNN}
\SetKwFunction{PredictDGCNN}{PredictDGCNN}
\SetKwFunction{ExtractFeatures}{ExtractFeatures}
\SetKwFunction{Concatenate}{Concatenate}
\SetKwFunction{Balance}{Balance}
\SetKwFunction{SelectBatch}{SelectBatch}

\SetKw{Break}{break}
\SetKw{Stop}{stop}
\SetKw{Input}{Input:}
\SetKw{Output}{Output:}

\SetAlgoLined
\Output Feasibility Prediction $\hat{\vec{y}}_{test}$  

\Input Local Search (\LS), Look-ahead Window ($W$), $r$, $L$, \DGCNN, $e$, $b$ 

\For{$l\leftarrow 1$ \KwTo $L$}{
 Generate $G_0$, 
 $i  \longleftarrow 0$\;
 \While{running time $\leq r$ }{
  \lIf{$G_{i}$ is feasible}{\Stop  
  }
   $G_{i+1}  \longleftarrow$ \LS{$G_{i}$}\;
   $i  \longleftarrow i+1$\;
  }
  $N_l \longleftarrow i$
  
  \For{$i \leftarrow 0$ \KwTo $N_l$}{
  $X_i \leftarrow \ExtractFeatures(G_i)$\;
  \lIf{$G_{min\{N_l, W+i\}}$ is feasible}{$y_i \leftarrow 1 $}
  \lElse{$y_i \leftarrow 0 $}
  }
  $\vec{X}_l$, $\vec{y}_l$  $\leftarrow$ \Concatenate{$X_0, ..., X_{N_l}$; $y_0, ..., y_{N_l}$ }\; 
}

$\{\vec{X}_{train}, \vec{y}_{train} \}_{j=1}^{m}$, $\{\vec{X}_{test}, \vec{y}_{test} \}_{j=1}^{m^\prime}$ $\leftarrow$ \Balance{$\vec{X}_1, ..., \vec{X}_{L}$; $\vec{y}_1, ..., \vec{y}_{L}$}\;

\While{number of epochs $\leq e$}{
   \For{$j \leftarrow 0$ \KwTo $ \lfloor{m/b}\rfloor $}{
 $\tilde{\vec{X}}$, $\tilde{\vec{y}}$ $\leftarrow $  \SelectBatch{$\vec{X}_{train},\vec{y}_{train}$}\;
 \TrainDGCNN{$\tilde{\vec{X}}$,$\tilde{\vec{y}}$}\;
 
}
}
$\hat{\vec{y}}_{test} \leftarrow$ \PredictDGCNN{$\vec{X}_{test}$}\;

\caption{General Graph Feasibility Classification}
\label{alg:DGCNN} 
\end{algorithm}

We denote the number of nodes in a graph $G_i$ as $n_{i}$, and extract $c$ node labels and temporal features from each graph $G_i$ to form an input matrix $X_i \in \mathbb{R}^{n_i \times c}$. Then, we collect $N_l + 1$ (including the initial solution) matrices $X_i$, to form tensors $\vec{X}_l$ where each row is a matrix $X_i$. Similarly, we collect classes $y_i$ to form  $ \vec{y}_l \in \mathbb{Z}_2^{(N_l + 1) \times 1}$. Then, we run $L$ Local Search procedures and concatenate instances $\vec{X}_l$ and $\vec{y}_l$ to form $\vec{X}$ with $N_L = \sum_{l=1}^{L} N_l + 1$ matrices $X_i$, and $\vec{y}$ with $N_L$ rows, each containing a binary class. Next, we separate the data between training $\{\vec{X}_{train}, \vec{y}_{train} \}_{j=1}^{m}$ and testing examples, $\{\vec{X}_{test}, \vec{y}_{test} \}_{j=1}^{m^\prime}$. However, to avoid biasing the DGCNN towards any given instance, we randomly select the same number of graphs from any given instance (LS run) $l$. We proceed to undersample the majority class until we get $m$ training and $m^\prime$ testing examples. During training, we randomly select $b$ training examples $X_i$ and $y_i$, in mini-batches of data and pass as inputs to the modified DGCNN algorithm. Finally, we train the network to classify between feasible and infeasible solutions until we observe a total number of epochs $e$. The proposed %version of the
feasibility classification algorithm is shown in Algorithm \ref{alg:DGCNN}.
\section{Results}
\label{sec:results}
In this section, we first present the experimental setups and the data used in the evaluations. We then evaluate the performance of the proposed method in predicting the feasibility of an initial solution. Next, we generalise the model to consider an arbitrary graph in the local search solution sequence. Also, we show that the proposed method can be used in combination with the local search to escape unpromising search areas. Finally, we estimate the difference in running time with and without the DGCNN predictions.
% We show how the proposed methodology can be used to predict the feasibility at the start of the LS run. 
\subsection{Setup of Experiments}

\subsubsection{Data Generation}  We generate data instances from the instance generator in the shunt plan simulator. The instance generator can be specified according to a set of input parameters based on the day-to-day schedule at the given service site. The most important parameters include: (1) number of train units, (2)  different train unit types and sub-types, (3) probability distributions of arrivals per unit type, and (4) set of service tasks including duration.
We generated 1,489 instances for training purposes and 1,143 instances for testing purposes, i.e. 2,632 instances in total, with varying initial random seeds. We ran the shunting plan simulator for the same amount of time (three days) to generate both training and testing instances, with the difference in numbers being the difficulty of solving the instances and machine running time. All instances were generated with 21 train units based on the ``average'' service site ``Kleine Binckhorst'' operated by NS. The amount of train units $\tau = 21$ has been purposely chosen as an increasing number of train units increases the difficulty in finding feasible solutions. As seen in previous works \cite{Dai2018,arno}, 21 trains yields a balanced yet challenging feasibility problem. Initial solutions were created for all instances and the LS procedure was applied to search for feasible solutions. The maximum running time allowed for LS is set to $r = 300$ seconds. We selected the remaining input parameters under ``normal operation" conditions, that is: two types of train units, two different sub-types (6 and 4 units), three service tasks (technical maintenance A/B and internal cleaning) and arrival/departure distributions matching real-world scenarios.

\begin{figure}[!ht]
\centering
\begin{floatrow}
  \ffigbox[\FBwidth]{\caption{Distribution and mean values of number iterations for feasible and infeasible solutions}\label{fig:distributions}}{%
    \includegraphics[width=.905\linewidth]{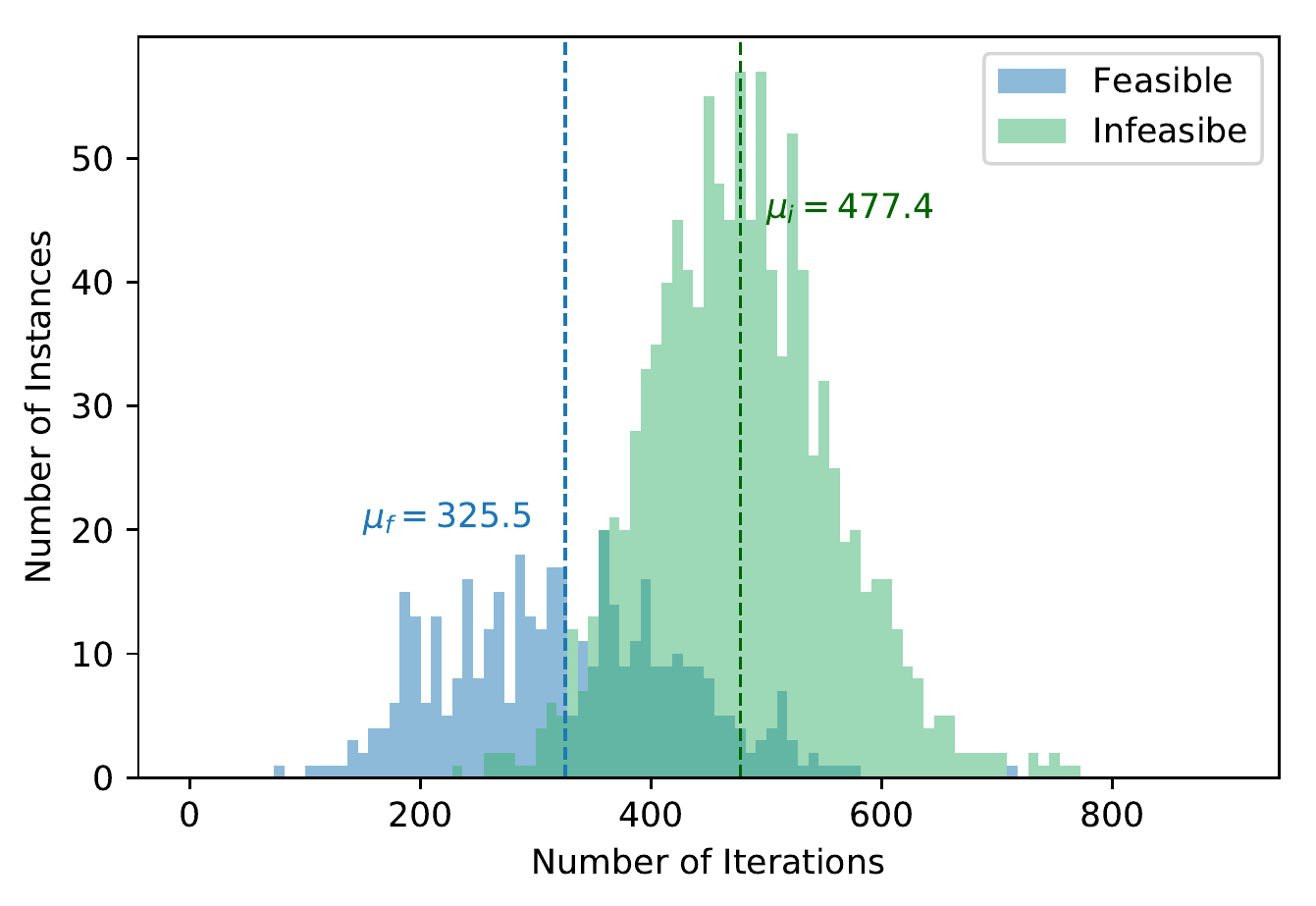}
  }
  \ffigbox[\FBwidth]{\caption{Objective function of the LS versus number of iterations for one feasible and infeasible example 
%   Random moves make it hard to predict feasibility given a plan.
  }\label{fig:cost}}{%
    \includegraphics[width=.92\linewidth]{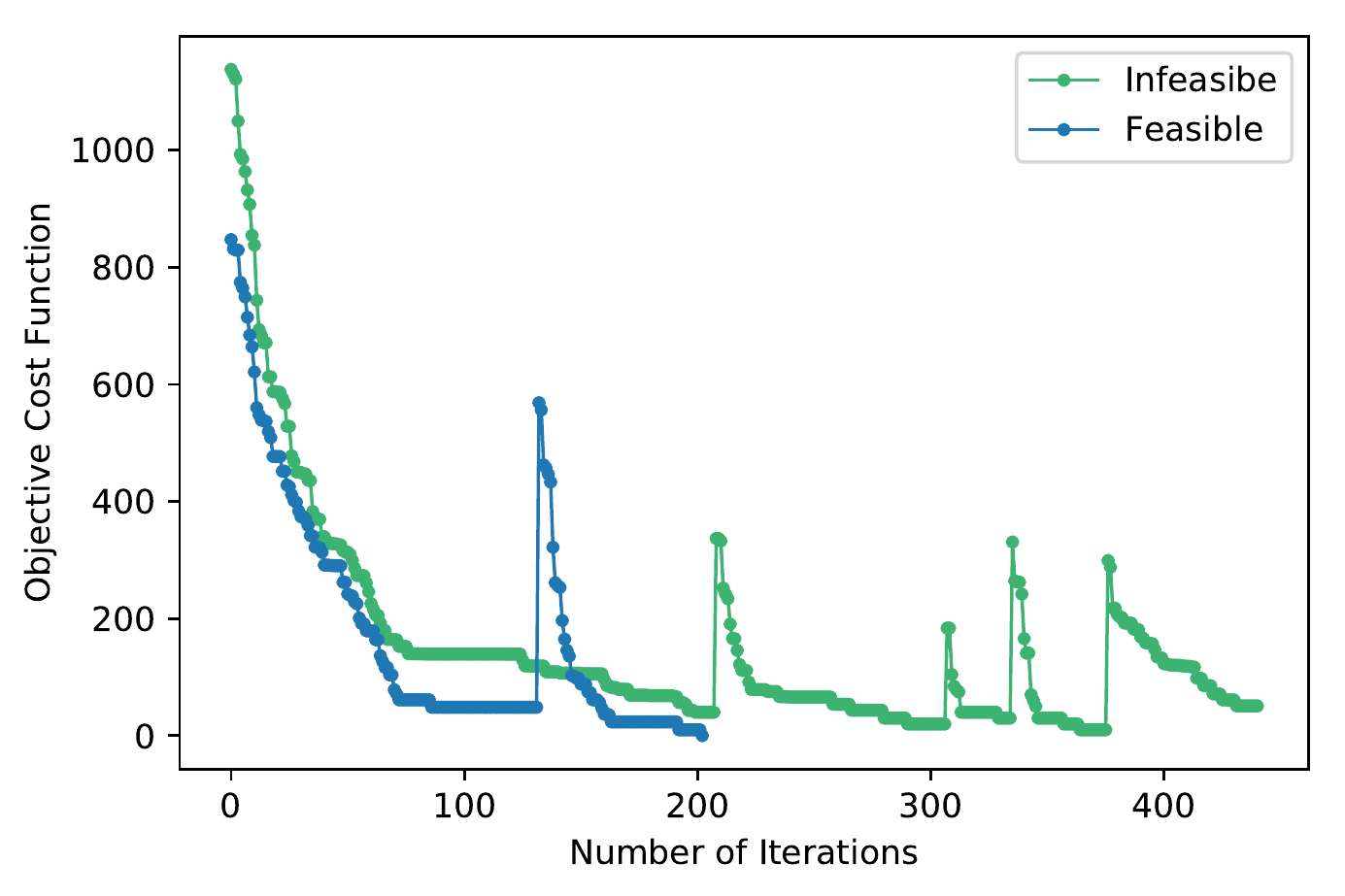}
  }
\end{floatrow}
\end{figure}
Among generated instances for training, 1,081 (72\%) are infeasible at the end of the LS run, while 408 (28\%) are considered feasible. In Figure \ref{fig:distributions}, we show the distribution of the number of iterations of the local search for feasible and infeasible instances. As can be noted from the distributions, our experimental setup is very challenging (imbalanced), i.e. it is hard for the LS to find feasible solutions in the maximum allowed running time. The main reasons for the difficulty in our dataset are related to the increased number of conflicts considered in the problem instance. Therefore, our model is faced with a much harder problem then previous classification tasks for the same number of train units \cite{Dai2018,arno}. Moreover, the LS also performs random moves as internal restarts when it cannot move to a better solution within a determined time limit (30 seconds). A typical plot of the cost function over time for a feasible and infeasible solution is shown in Figure \ref{fig:cost}. This stochastic behaviour of the search procedure makes it specially challenging to predict feasibility even when solution are ``close'' to feasibility.

\subsubsection{Experimental Settings} To accommodate for the imbalance between the classes, we undersample the majority class until we achieve a 50\% of the examples for each class. We perform 5-fold cross validation (4 folds for training and 1 fold for cross-validation) using one training fold for manual hyperparameter search. We report the results on the testing graphs (1,143 instances). Moreover, the DGCNN is implemented with graph convolutions with channels:  $32\times32\times32\times1$; unifying nodes in graph $k$: 0.6; learning rate: $1\mathrm{e}{-5}$; number of training epochs: 200 and batch size: 50. The remaining layers consist of two 1-D convolutional and MaxPooling layers and one dense layer implemented as described in the original paper. The dense layer has 128 hidden units followed by a $softmax$ layer as output. A dropout rate of 0.5 is used after the dense layer. The hyperbolic tangent function ($tanh$) is used in graph convolution layers, and rectified linear units ($ReLU$) in the remaining layers. Stochastic Gradient Descent (SGD) with the Adam updating rule \cite{kingma2014adam} were used for optimisation.

All our experiments were conducted on an Intel(R) Core(TM) i5-7200U 2.50GHz CPU, 8GB RAM and Nvidia GTX 1070 GPU. Implementation was done in C\# for the LS procedure and in Python (3.7.1). The PyTorch \cite{paszke2017automatic} library (0.2.2) and a modified implementation of the DGCNN \cite{Zhang2018} were used for training the neural network architecture.

\subsection{Initial Solution Classification Performance}
\label{subsec:initialclass}
In this section, we evaluate the classification performance of the graph classification models (Algorithm \ref{alg:DGCNN}) trained under different settings. We consider the following settings:

\begin{itemize}
    \item DGCNN-IS: A DGCNN using only node labels and initial solutions during training \cite{arno}  (\textit{$X_0$ with only node labels}).
    \item DGCNN-IS-T: A DGCNN with additional temporal features using only the initial solutions for training (\textit{$X_0$ with additional temporal features}).
    \item DGCNN-MS-T: A DGCNN with additional node temporal features using the first 10\% of the graphs in each instance of a LS run for training. (\textit{$X_0$--$X_{\lfloor 0.1 \cdot N_l \rfloor}$ in $\vec{X}_{train}$, $X_i$ with additional temporal features})
\end{itemize}

All models consider as class labels $y_i$ the final feasibility status at the end of an LS run ($W=N_l$). We perform inference on the initial solutions $G_0$ on the test dataset. The classification results, including \textit{accuracy} (ACC), \textit{true positive rate} (TPR) and \textit{true negative rate} (TNR) for the prediction models trained under different settings are displayed in Table \ref{tab:initial_acc}. All results displayed are calculated on balanced test instances.

\begin{table}[!ht]
\setlength{\tabcolsep}{10pt}
\caption{Comparison between the proposed models. Adding time features improves performance by 10\%.}
\label{tab:initial_acc}
\centering
\resizebox{0.8\columnwidth}{!}{%
\begin{tabular}{lccc}
\hline
\textbf{Method} & DGCNN-IS \cite{arno}          & DGCNN-IS-T                & DGCNN-MS-T                \\ \hline
ACC(\%)             & \small54.10 $\pm$ 2.37  & \textbf{60.01 $\pm$ 1.67} & 59.96 $\pm$ 1.68          \\
TPR(\%)               & 52.93 $\pm$ 11.47 & 58.25 $\pm$ 9.10          & \textbf{82.08 $\pm$ 2.53} \\
TNR(\%)               & 55.28 $\pm$ 10.27 & \textbf{60.38 $\pm$ 7.37} & 37.86 $\pm$ 4.99          \\ \hline
\end{tabular}
}
\end{table}

Results show that temporal features add important information for the classification of feasible instances. The model without time features (DGCNN-IS) shows a drop in performance in comparison to the results presented in \cite{arno}. This is due to more complicated instances being used, disallowing certain moves in the shunting yard. The DGCNN-IS-T achieves an improved performance of approximately 10\% when compared to the DGCNN-IS model. Lastly, the DGCNN-MS-T also shows similar accuracy performance with the model trained only on the initial solution. However, it has a much higher TPR than the other models, showing that the model is biased towards predicting positive (feasible) classes, but worse at capturing infeasible instances (TNR). These results motivate us to consider a variant of the the prediction model that can be used alongside the local search procedure.

\subsection{Evaluation in Combination with Local Search}
In this section we consider a variant of the prediction model of the previous subsection. We generalise the training of the DGCNN and show how to combine the prediction with the LS in to improve decision-making.

In the models of subsection \ref{subsec:initialclass}, the feasibility label at the end of a LS run was used for training the DGCNN. This can be generalised by using the concept of \emph{time-window look-ahead}: instead of looking at the label at the very end of a LS run, we look at the label of the graph that is \( W \) solutions ahead of the current one. We refer to this specification as DGCNN-MS-T-W (DGCNN using multiple solutions including temporal features looking ahead \( W \) graphs). The intuition behind this specification is to predict or score whether the graph \( W \) iterations in the future is similar to graphs that are within \( W \) iterations of being feasible. Here \( W \) controls how eager the decision-maker is to learn about feasibility (by looking ahead during a run of the LS).  

It is interesting to look at the predictions of the DGCNN-MS-T-W model over various runs of the LS. For each iteration \( i \)  in a run of LS, the associated graph \( G_{i} \) can be fed to DGCNN-MS-T-W and results in a predicted score  \( \phi(G_{i}) \). We note that the values of $\phi(G_{i})$ can vary according to the problem instance considered (e.g. varying $\tau$). Figure \ref{fig:score} shows the averaged (moving average over the last 10 iterations) scores \( \phi(G_{i}) \) over the cross-validation runs of the LS for both feasible and infeasible instances. From Figure \ref{fig:score} we observe the following patterns: the feasibility scores are lower for infeasible solutions when compared to feasible solutions. Moreover, scores decay over time for observed infeasible solutions, showing that the model is capturing the look-ahead prediction function. In the figure, there is no clear constant threshold value that could be derived for all iterations. Therefore, we need to look at other metrics to define an appropriate value. 

The observed patterns can be used to design a simple policy that combines a trained DGCNN-MS-T-W model with the LS. For example, given an instance of TUSP, one can use the following threshold policy:
\begin{enumerate}
    \item Start the LS with the initial solution and run it for \( K \) iterations.
    \item For each iteration $i$ where \(  0 \leq i \leq K \), we pass the current solution \( G_i \) to DGCNN-MS-T-W to get a feasibility score \( \phi(G_i) \). 
    % \item If the feasibility score \( f_{i}(G) \) exceeds  some threshold \( 0 < \alpha_{F} < 1\), then we stop the LS and classify the instance as feasible. 
    \item If an arbitrary function $g$ of the feasibility scores \( \phi(G_i) \) seen up to iteration $K$ falls below some threshold \( 0 < \alpha_{IF} < 1\), we stop the LS and classify the instance as infeasible. 
\end{enumerate}
We show the general form of the policy in Algorithm \ref{alg:policy}.
\begin{algorithm}[ht]
\SetKwFunction{LS}{LS}
\SetKwFunction{DGCNNMSTWNAME}{DGCNN-MS-T-W}
\SetKwFunction{DGCNNMSTW}{PredictDGCNN-MS-T-W}
\SetKwFunction{TrainDGCNN}{TrainDGCNN}
\SetKwFunction{PredictDGCNN}{PredictDGCNN}
\SetKwFunction{ExtractFeatures}{ExtractFeatures}
\SetKwFunction{Concatenate}{Concatenate}
\SetKwFunction{Balance}{Balance}
\SetKwFunction{SelectBatch}{SelectBatch}

\SetKw{Break}{break}
\SetKw{Stop}{stop}
\SetKw{Input}{Input:}
\SetKw{Output}{Output:}

\SetAlgoLined
\Output Feasibility of $l$

\Input Local Search (\LS), Trained \DGCNNMSTWNAME, $r$, Instance: $l$

Generate $G_0$, 
$i  \longleftarrow 0$\;

 \While{running time $\leq r$ }{
   \If{$G_{i}$ is feasible}{$l$ $\longleftarrow$ feasible\; \Stop\;}
   \If{$i \leq K$}{
      $\phi(G_i)$  $\longleftarrow$ \DGCNNMSTW{$G_i$}\;
      \If{$g(\phi(G_0),...,\phi(G_i)) \leq \alpha_{IF}$ }{ 
    $l \longleftarrow$ infeasible\;
    \Stop\;
    }  
    }
    \Else{
    $G_{i+1}  \longleftarrow$ \LS{$G_{i}$}\;
    $i  \longleftarrow i+1$\;
    }
     
  }
\caption{Threshold Policy for the TUSP Local Search}
\label{alg:policy} 
\end{algorithm}

The values of \( \alpha_{IF} \) can, for example, be obtained from the analysis of Figure \ref{fig:score} and can vary for different difficulties of the TUSP. While the value $K$ can be estimated empirically using the available training data. For example, we could select $K$ by considering the expected \textit{number of iterations} until feasibility or based on the expected \textit{running time} until a certain iteration. The main motivation for considering such a procedure is the expected reduction in computation time. If we have a reasonable prediction model, then computational resources are used more efficiently because we only continue the LS procedure if we are highly uncertain about (in-)feasibility. 
\subsubsection{Performance Gains in Combination with LS}  
To quantify the added value of a procedure like the threshold policy, we consider the following performance indicators: (1) Classification metrics such as accuracy, false positive rate, false negative rate, true positive rate and true negative rate. These metrics are important as they show how often the threshold policy comes to the same conclusion as LS. (2) Saved computational time. In those cases that the threshold policy and the LS come to the same conclusion, we can look at the running time that is saved by the threshold policy. 
We define the constants $K = 200$ iterations since at that point the LS has spent on average 80 seconds (26\% of the total running time) looking for a solution. For the purpose of more stability in the predictions, we define $g$ to be the average score between iterations $0$ an $K$. That is, we consider as a decision point iteration $K$. However, other functions are possible, for example, one could consider each $\phi(G_i)$, $0 \leq i \leq K$ as a decision point. In our tests, using each $\phi(G_i)$ resulted in too many disagreements between predictions and feasibility, leading to wasted computational time. Moreover, since we expect that feasible solutions will be found on average in 325 iterations, we define our look ahead window $W = 150$ to accommodate that interval with $K + W$.
\begin{figure}
\begin{floatrow}
\ffigbox{\caption{Moving average over the last 10 iterations of the feasibility scores $\phi(G_i)$ averaged over cross-validation runs of the LS procedure}\label{fig:score}}{%
  \resizebox{1\columnwidth}{!}{%
    \includegraphics{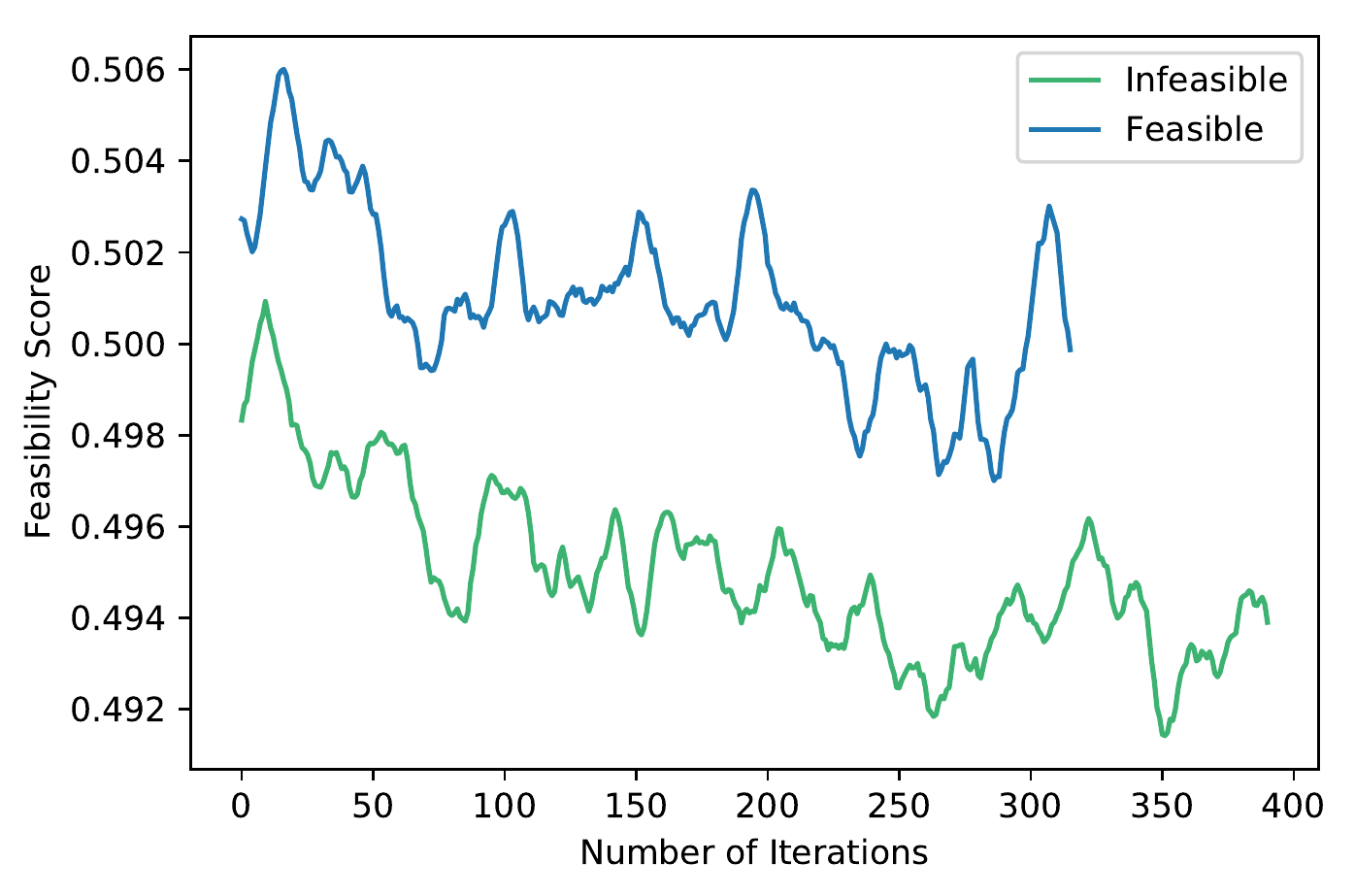}
    }%
}{%
}
\capbtabbox{\caption{Confusion matrix of one of the folds for the final classification of DGCNN-MS-T-W, with $W= 150$}\label{tab:conf_mat}}{%
\renewcommand\arraystretch{1.5}
\setlength\tabcolsep{0pt}
\scalebox{0.6}{% 
    
    \begin{tabular}{c >{\bfseries}r @{\hspace{0.7em}}c @{\hspace{0.4em}}c @{\hspace{0.7em}}l}
      \multirow{13}{*}{\rotatebox{90}{\parbox{1.1cm}{\bfseries\centering Actual\\ Value}}} & 
        & \multicolumn{2}{c}{\bfseries Prediction Outcome} & \\
      & & \bfseries 0: infeasible & \bfseries 1: feasible & \bfseries \begin{tabular}[c]{@{}c@{}}Correct\\ Incorrect\end{tabular}  \\
      & 0 & \MyBox{543}{33.9\%} & \MyBox{250}{15.6\%} &  \begin{tabular}[l]{@{}l@{}}68.4\%\\ 31.6\%\end{tabular} \\[2.4em]
      & 1 & \MyBox{284}{17.7\%} & \MyBox{523}{32.7\%} & \begin{tabular}[l]{@{}l@{}}64.8\%\\ 35.2\%\end{tabular}  \\
      & \begin{tabular}[c]{@{}c@{}}Correct\\ Incorrect\end{tabular}  & \begin{tabular}[l]{@{}l@{}}65.6\%\\ 34.4\%\end{tabular} & \begin{tabular}[l]{@{}l@{}}67.6\%\\ 32.4\%\end{tabular} & \begin{tabular}[l]{@{}l@{}}\cellcolor{gray!25}\textbf{66.6\%}\\ 33.4\%\end{tabular} 
    \end{tabular} 
}
}
\end{floatrow}
\end{figure}

In Table \ref{tab:conf_mat}, we show the results of the confusion matrix coming from one of the folds in our experiments. The DGCNN-MS-T-W model achieves the best accuracy of all attempted models with accuracy: 64.0\% $\pm$ 1.07 over all 5-folds. However, this result would not be beneficial if the policy to halt the LS does not lead to saved computational time. To maintain consistency with our previous models, we define a threshold $\alpha_{IF} = 0.5$ to calculate the new running times considering the new policy.
We compute the difference between the expected running time of the LS before and after the policy based on the DGCNN. We use the confusion matrix percentages from Table \ref{tab:conf_mat}, the average running times (feasible: 157 seconds, infeasible: 300 seconds) weighted by the original imbalanced data to compute estimates based on a Markov chain. Such Markov chain arises as even after stopping the LS, we are still uncertain about the feasibility of the next time we run the LS for the same instance.

After computing the running times, we achieve a 8\% reduction for a single instance, which can account for roughly 20 hours in total real time. We point out that the proposed policy does not halt the LS when it is ``certain'' about feasibility. A change in the policy to consider such cases, can yield gains up to 30\% in running times with the counterpart of losing some feasibility certainty. Lastly, we point out that the extra burden to calculate the scores $\phi(G_i)$ for the solutions only adds little computational time as it only requires scoring a small number of graphs ($K=200$) during each LS run. 

\section{Discussion and Conclusion}

We studied the Train Unit Shunting Problem that is faced by NS. This problem involves matching arriving train units to service tasks and determining the schedule for departing trains. The TUSP is an important problem as it is used to determine the capacity of shunting yards and arises as a sub-problem of more general scheduling and planning problems. As the TUSP is a complex problem, NS currently uses a local search (LS) heuristic to determine if an instance of TUSP has a feasible solution. In the LS, solutions are represented as an activity graph and the LS takes as input an initial solution produced by an initial solution generator.

We showed how a machine learning approach can be used in combination with a local search heuristic to improve decision-making. First, we focused on predicting feasibility of an instance of TUSP at the start of a run of the LS. A Deep Graph Convolutional Neural Network is used as a prediction model to determine feasibility of a shunting plan. We employed different training strategies such as (i) training on the initial solution; (ii) training on the initial solution including temporal features; (iii) training on multiple solutions and including temporal features. We showed that training based on (ii) achieved an accuracy of 60\%, a 10\% relative improvement over the baseline (i). Our second contribution expands the original models to account for arbitrary graphs during an LS run. We control the eagerness to find a feasible solution by setting the labels over a number of iterations ahead. We show that the best model achieves the accuracy of 64\%.  We also study how such model can be used in combination with the LS. We evaluate the effect of a policy using the proposed models and show that it can lead to reduced running times.

An interesting direction for future work is to consider other aspects in addition to feasibility in a multi-task learning approach. For example, shunting plans with a low number of crossings are generally preferred by decision-makers. Moreover, in our current work we only decide whether to keep running the LS or to stop its execution.  Another direction is to design machine learning algorithms that interact directly with the LS operators and can select the most suitable operators given a certain plan.   

\section*{Acknowledgements}
\noindent The work is partially supported by the NWO funded project Real-time data-driven maintenance logistics (project number: 628.009.012).

%\bibliographystyle{splncs04}
%\bibliography{cite_peipei}

\begin{thebibliography}{10}
\providecommand{\url}[1]{\texttt{#1}}
\providecommand{\urlprefix}{URL }
\providecommand{\doi}[1]{https://doi.org/#1}

\bibitem{Spoorenplan}
Sporenplanonline. \url{http://www.sporenplan.nl/}, accessed: 2019-03-20

\bibitem{aggarwal2014evolutionary}
Aggarwal, C., Subbian, K.: Evolutionary network analysis: A survey. ACM
  Computing Surveys (CSUR)  \textbf{47}(1), ~10 (2014)

\bibitem{akoglu2015graph}
Akoglu, L., Tong, H., Koutra, D.: Graph based anomaly detection and
  description: a survey. Data mining and knowledge discovery  \textbf{29}(3),
  626--688 (2015)

\bibitem{bonner2016deep}
Bonner, S., Brennan, J., Theodoropoulos, G., Kureshi, I., McGough, A.S.: Deep
  topology classification: A new approach for massive graph classification. In:
  2016 IEEE International Conference on Big Data (Big Data). pp. 3290--3297
  (2016)

\bibitem{Vandenbroek2016}
van~den Broek, R.: Train Shunting and Service Scheduling: an integrated local
  search approach. Master's thesis, Utrecht University (2016)

\bibitem{Dai2018}
Dai, L.: A machine learning approach for optimization in railway planning.
  Master's thesis, Delft University of Technology (March 2018)

\bibitem{freling2005shunting}
Freling, R., Lentink, R.M., Kroon, L.G., Huisman, D.: Shunting of passenger
  train units in a railway station. Transportation Science  \textbf{39}(2),
  261--272 (2005)

\bibitem{haahr2017integrating}
Haahr, J., Lusby, R.M.: Integrating rolling stock scheduling with train unit
  shunting. European Journal of Operational Research  \textbf{259}(2),
  452--468 (2017)

\bibitem{haahr2017optimization}
Haahr, J.T., Lusby, R.M., Wagenaar, J.C.: Optimization methods for the train
  unit shunting problem. European Journal of Operational Research
  \textbf{262}(3),  981--995 (2017)

\bibitem{hopcroft1973n}
Hopcroft, J.E., Karp, R.M.: An n\^{}5/2 algorithm for maximum matchings in
  bipartite graphs. SIAM Journal on computing  \textbf{2}(4),  225--231 (1973)

\bibitem{Khalil_NIPS2017_7214}
Khalil, E., Dai, H., Zhang, Y., Dilkina, B., Song, L.: Learning combinatorial
  optimization algorithms over graphs. In: Advances in Neural Information
  Processing Systems. pp. 6348--6358 (2017)

\bibitem{kingma2014adam}
Kingma, D.P., Ba, J.: Adam: A method for stochastic optimization. arXiv
  preprint arXiv:1412.6980  (2014)

\bibitem{Kipf2016}
Kipf, T., Welling, M.: Semi-supervised classification with graph convolutional
  networks. CoRR  (2016)

\bibitem{kroon2008shunting}
Kroon, L.G., Lentink, R.M., Schrijver, A.: Shunting of passenger train units:
  an integrated approach. Transportation Science  \textbf{42}(4),  436--449
  (2008)

\bibitem{Lombardi2018}
Lombardi, M., Milano, M.: Boosting combinatorial problem modeling with machine
  learning. In: IJCAI-18. pp. 5472--5478 (2018)

\bibitem{neumann2016propagation}
Neumann, M., Garnett, R., Bauckhage, C., Kersting, K.: Propagation kernels:
  efficient graph kernels from propagated information. Machine Learning
  \textbf{102}(2),  209--245 (2016)

\bibitem{Niepert2016}
Niepert, M., Ahmed, M., Kutzkov, K.: Learning convolutional neural networks for
  graphs. CoRR  (2016)

\bibitem{paszke2017automatic}
Paszke, A., Gross, S., Chintala, S., Chanan, G., Yang, E., DeVito, Z., Lin, Z.,
  Desmaison, A., Antiga, L., Lerer, A.: Automatic differentiation in pytorch
  (2017)

\bibitem{peer2018shunting}
Peer, E., Menkovski, V., Zhang, Y., Lee, W.J.: Shunting trains with deep
  reinforcement learning. In: 2018 IEEE International Conference on Systems,
  Man, and Cybernetics (SMC). pp. 3063--3068. IEEE (2018)

\bibitem{shervashidze2011weisfeiler}
Shervashidze, N., Schweitzer, P., Leeuwen, E.J.v., Mehlhorn, K., Borgwardt,
  K.M.: Weisfeiler-lehman graph kernels. Journal of Machine Learning Research
  \textbf{12}(Sep),  2539--2561 (2011)

\bibitem{arno}
van~de Ven., A., Zhang., Y., Lee., W., Eshuis., R., Wilbik., A.: Determining
  capacity of shunting yards by combining graph classification with local
  search. In: ICAART - Volume 2. pp. 285--293 (2019)

\bibitem{verwer2017auction}
Verwer, S., Zhang, Y., Ye, Q.C.: Auction optimization using regression trees
  and linear models as integer programs. Artificial Intelligence  \textbf{244},
   368--395 (2017)

\bibitem{Zhang2018}
Zhang, M., Cui, Z., Neumann, M., Chen, Y.: An end-to-end deep learning
  architecture for graph classification. In: AAAI-18. pp. 4438--4445 (2018)

\end{thebibliography}

\end{document}